\documentclass[11pt]{article}

\usepackage[preprint]{acl}

\usepackage{times}
\usepackage{latexsym}

\usepackage[T1]{fontenc}

\usepackage[utf8]{inputenc}

\usepackage{microtype}

\usepackage{inconsolata}

\usepackage{graphicx}

\usepackage{times}
\usepackage{latexsym}
\usepackage[T1]{fontenc}
\usepackage[utf8]{inputenc}
\usepackage{microtype}
\usepackage{inconsolata}
\usepackage{graphicx}

\usepackage{amsmath}
\usepackage{amssymb}
\usepackage{booktabs}
\usepackage{algorithm}
\usepackage{algpseudocode}
\usepackage{array}
\usepackage{multirow}
\usepackage{tabularx}
\usepackage{bbm}

\usepackage{placeins}

\title{Risk-Constrained Freshness-Aware Semantic Caching for Open-Web Retrieval-Augmented LLMs}

\author{
  \textbf{Muhammad Mansoor\textsuperscript{1}},
  \textbf{Tahir Ahmad\textsuperscript{1}},
  \textbf{Yeo-Chan Yoon\textsuperscript{1}\thanks{Corresponding author.}}
\\
\\
  \textsuperscript{1}Jeju National University,
  Jeju, Republic of Korea
\\
  \small{
    \texttt{mmansoor7755@stu.jejunu.ac.kr},
    \texttt{tahirahmad@stu.jejunu.ac.kr},
    \texttt{ycyoon@jejunu.ac.kr}
  }
\\
  \small{
    \textbf{Correspondence:} \href{mailto:ycyoon@jejunu.ac.kr}{ycyoon@jejunu.ac.kr}
  }
}

\begin{document}
\maketitle
\begin{abstract}
Semantic caching reduces the latency and cost of retrieval-augmented generation (RAG) by serving cached answers to semantically similar queries, but most existing methods do not model the time-varying freshness of open-web evidence. We present FreshCache, a three-tier semantic cache that treats cache reuse as a risk-constrained temporal inference problem: before approving a cache hit, FreshCache estimates the probability that the cached result is stale using a fitted exponential decay model enhanced by a learned MLP, and approves reuse only when that probability falls below a per-tier error budget across answers ($\varepsilon = 0.10$), URL lists ($\varepsilon = 0.20$), and page content ($\varepsilon = 0.35$). This allows the system to degrade gracefully as entries age rather than forcing a binary choice between a stale hit and a full pipeline execution. We introduce FreshCache-Bench, a benchmark of 8,072 base queries across five freshness classes with ground truth staleness labels drawn from real web snapshots at 1, 12, 24 hours, and 7 days after a baseline crawl, expanded to 31,201 queries via paraphrase generation. At the 24-hour evaluation window, FreshCache\_MLP achieves 97\% search API savings at 0.1\% hash-based stale error, and an LLM-judge evaluation on 396 confirmed change pairs shows that only 34.3\% of detected content changes actually affect answer correctness, placing true answer-affecting stale error at approximately 0.034\%. The rule-based FreshCache achieves 98\% search savings at 3.3\% stale error under a temporal holdout calibration, outperforming SemanticTTL (14.9\% stale, 72\% saved), vCache (7.2\% stale, 47\% saved), and SCALM (5.2\% stale, 96\% saved). Ablations show the temporal risk gate accounts for an 11.6 point reduction in stale error over similarity-only reuse, and the learned MLP reduces stale error a further 3.2 points over the rule-based model.
\end{abstract}

\section{Introduction}
\label{sec:intro}

Retrieval-augmented generation grounds LLM outputs in external knowledge~\cite{lewis2020rag, gao2023survey}. In open-web RAG systems, each query triggers a search API call, HTTP fetches of the returned pages, and an LLM call that synthesizes an answer from the retrieved content. Latency routinely exceeds three seconds per query, and at scale the API and compute costs become substantial.

Semantic caching reduces this overhead~\cite{bang2023gptcache, zhu2024semsimilarity}: if an incoming query is semantically similar to a previously answered one, the system serves the cached response instead of repeating the pipeline. Prior systems report search cost reductions of more than half on workloads with recurring query patterns~\cite{li2024scalm, schroeder2025vcache}.

The problem is that existing semantic caches largely ignore a basic property of open-web content: it changes over time. A cached answer that was accurate when stored may be wrong by the time it is served, and how quickly this happens varies enormously by query type. A country capital can be reused for months; a live sports score is stale within minutes. Fixed TTL systems face an uncomfortable trade-off: too short and cache efficiency is lost, too long and stale answers erode user trust.

We argue that reusing a cached answer is fundamentally a risk-constrained temporal inference problem. Before serving a cached result, the right question is not only whether the query is similar enough, but also, given the content type, the entry's age, and the retrieval tier, what is the probability that this response is stale, and is that probability within an acceptable error budget. Framing caching as staleness risk estimation rather than similarity thresholding is the central idea behind FreshCache.

FreshCache is a three-tier semantic cache built for open-web RAG. The tiers mirror the pipeline's natural stages: final answers (L1), retrieved URL lists (L2), and raw page content (L3). Each tier carries a configurable error budget. Before serving from any tier, FreshCache estimates staleness probability using a calibrated risk model and approves reuse only when it falls within budget. When L1 risk is too high, the system degrades to L2, saving the search cost while generating a fresh answer from re-fetched content; when even L2 is too risky, conditional GET validation at L3 confirms whether pages changed before an unconditional re-fetch. FreshCache therefore always operates at the highest safe tier rather than making a binary cache-or-miss decision.

We evaluate FreshCache on FreshCache-Bench, 8,072 real web queries across five freshness classes (timeless, slow, medium, fast, real time), re-fetched at one, twelve, and twenty four hours and again at seven days to produce ground truth staleness labels, expanded to 31,201 queries via paraphrasing. We compare against three semantic caching baselines (SemanticTTL~\cite{bang2023gptcache}, vCache~\cite{schroeder2025vcache}, SCALM~\cite{li2024scalm}), two class aware TTL baselines, a tiered fixed TTL ablation, and an oracle upper bound. All experiments use the BAAI/bge-m3 multilingual encoder.

At the 24-hour window, FreshCache\_MLP reaches 0.1\% stale error and 97\% search savings, beating every baseline on both axes. The gap between the rule-based and MLP variants is explained by within-class discrimination: the MLP identifies stable queries inside volatile classes and approves L1 reuse for them, recovering savings the class-level rule is too conservative to capture.

Our contributions are: (1) a formulation of open-web RAG caching as a risk-constrained temporal reuse problem, with an empirical case for why similarity alone is insufficient; (2) FreshCache, a three-tier architecture with per-tier staleness budgets, along with a negative finding that lightweight content validation is largely ineffective on the open web, which motivates risk-gated L1 and L2 reuse as the primary source of savings; (3) FreshCache-Bench, a temporally grounded benchmark with real snapshot based staleness labels; (4) a comprehensive evaluation against five baselines and an oracle, showing the risk gate and learned model each contribute a measurable reduction in stale error; and (5) validation of the hash-based stale proxy against an LLM judge, and a live-web microbenchmark confirming the simulation's latency assumptions are conservative. Full details of several supporting analyses appear in the appendix, cited at the relevant points below.

\section{Related Work}
\label{sec:related}

Retrieval-augmented generation was introduced by Lewis et al.~\cite{lewis2020rag}, with Dense Passage Retrieval~\cite{karpukhin2020dpr} providing a standard dense retrieval backbone. Nakano et al.~\cite{nakano2021webgpt} extended this to live web browsing, and Lazaridou et al.~\cite{lazaridou2022internet} and Mallen et al.~\cite{mallen2023trust} show that fresh retrieval substantially improves accuracy on time-sensitive and long-tail queries. This line of work focuses on what is retrieved and how quickly; FreshCache instead asks whether a previous retrieval result can be safely reused.

Semantic caching for LLM services was popularized by GPTCache~\cite{bang2023gptcache}, formalized by Zhu and Zhu~\cite{zhu2024semsimilarity}, and extended by SCALM~\cite{li2024scalm} through hierarchical clustering and by vCache~\cite{schroeder2025vcache} through per-embedding threshold learning with formal error guarantees. These systems model semantic match quality and cache-hit correctness, but none explicitly estimate open-web evidence staleness as a time dependent risk; a semantically correct hit can still return stale information if the underlying page has changed. FreshCache keeps the semantic lookup machinery from GPTCache and SCALM and the error budget framing from vCache, and adds tier-specific staleness-risk estimation as an independent gate on top of semantic relevance.

Wang et al.~\cite{wang2025categoryaware} vary similarity thresholds and TTLs per query category, but assign a static per-category TTL on a single-tier cache. FreshCache instead estimates a calibrated probability at query time conditioned on freshness class, domain volatility, and query features, and applies a distinct risk budget across three tiers rather than one. RAGCache~\cite{jin2024ragcache}, CacheBlend~\cite{yao2025cacheblend}, and CAG~\cite{chan2025cag} accelerate generation by caching KV representations of a static corpus and never ask whether cached content remains correct; these are complementary to FreshCache rather than competing with it. On the temporal knowledge side, Dhingra et al.~\cite{dhingra2022time}, Tan et al.~\cite{tan2023temporal}, and Kasai et al.~\cite{kasai2023realtime} document systematic LLM failures on time-scoped facts, and FreshLLMs~\cite{vu2024freshllms} proposes a five-class volatility taxonomy that directly motivates our freshness classes, though it re-fetches evidence unconditionally rather than gating reuse. Web change dynamics studied by Fetterly et al.~\cite{fetterly2003evolution}, Cho and Garcia-Molina~\cite{cho2003frequency}, and Olston and Pandey~\cite{olston2008recrawl} inform our decay model. All embeddings use BAAI/bge-m3~\cite{chen2024bgem3}, a multilingual encoder building on Sentence-BERT~\cite{reimers2019sbert}, chosen to support both English and Korean queries in our benchmark. Table~\ref{tab:related} in Appendix~\ref{app:related} summarizes how FreshCache differs from each of these systems along six design dimensions.

\section{Methodology}
\label{sec:system}

\begin{figure*}[!t]
    \centering
    \includegraphics[width=1.0\textwidth, height=0.35\textheight]{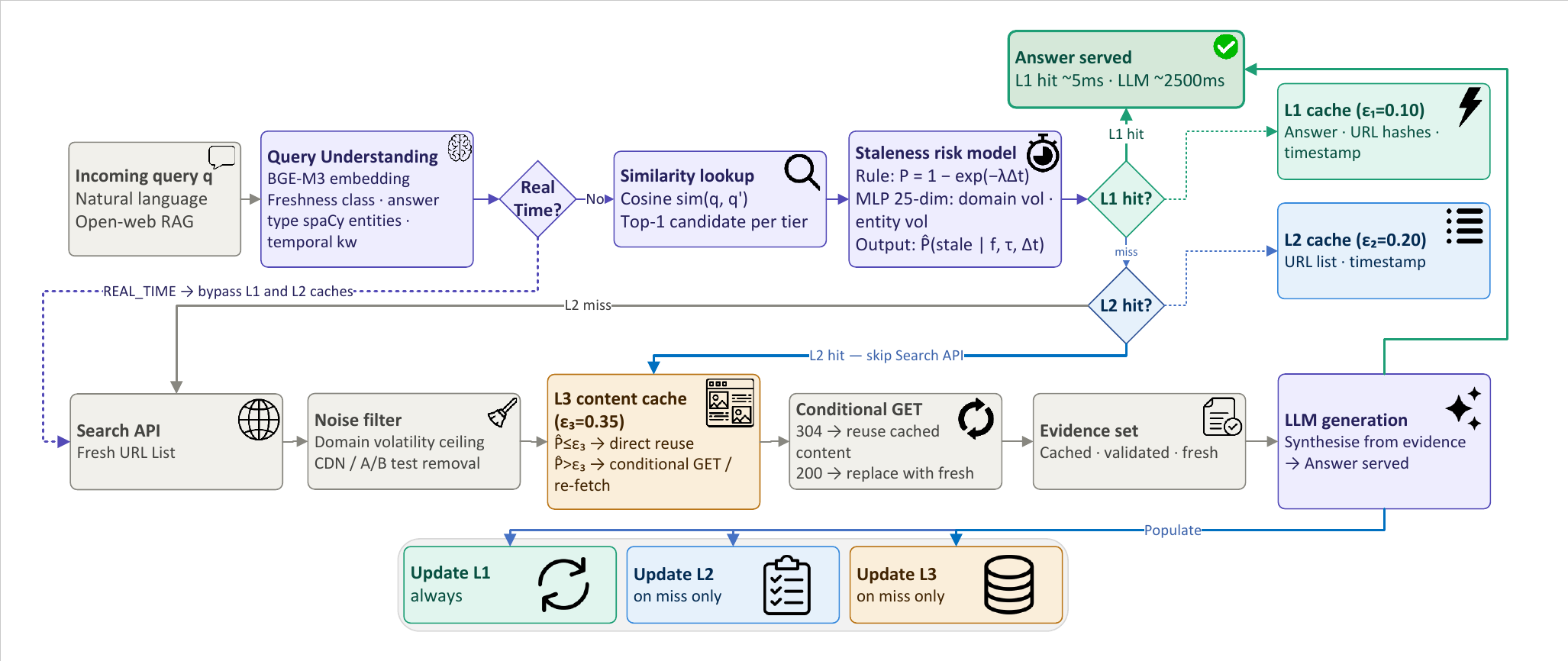}
    \caption{FreshCache system architecture showing semantic lookup, risk-gated cache reuse, and fallback execution across the L1, L2, and L3 cache tiers.}
    \label{fig:architecture}
\end{figure*}

\subsection{Problem Formulation}
\label{sec:problem}

An open-web RAG pipeline processes queries $q_1, q_2, \ldots$. For each query $q$, the pipeline runs a search call returning URLs $U(q)$, an HTTP fetch producing content $C(q)$, and an LLM call producing an answer $A(q)$. A semantic cache stores records $(q', U', C', A', t)$, where $t$ is the population time. We define a cached entry as stale if any URL in $U'$ has changed since $t$:
\begin{equation}
  \text{stale}(q', t, t') = \max_{u \in U'} \; \delta(u, t, t')
  \label{eq:stale}
\end{equation}
where $\delta$ is a binary change indicator. Given an error budget $\varepsilon_\tau$ per tier, the reuse decision is:
\begin{equation}
\begin{aligned}
\tau^*(q) &= \arg\max_{\tau \in \{L1,L2,L3\}}
               \mathbbm{1}\!\left[C_\tau(q,q',t)\right],\\[4pt]
C_\tau(q,q',t)
  &:\; \text{sim}(q,q') \ge \theta_\tau \\
  &\quad\wedge\;
    \hat{P}(\text{stale} \mid q',\tau,\,t_{\text{now}}{-}t)
    \le \varepsilon_\tau .
\end{aligned}
\label{eq:decision}
\end{equation}
Reuse is therefore a conjunction of a semantic condition and a temporal condition; satisfying similarity alone is not sufficient.

\subsection{Freshness Classification}
\label{sec:classification}

We partition queries into five freshness classes, following the taxonomy in FreshLLMs~\cite{vu2024freshllms}: TIMELESS (historical and definitional facts), SLOW (corporate and policy facts that change over weeks to months), MEDIUM (financial and news facts that may change within days), FAST (live scores and prices that change within hours), and REAL\_TIME (traffic, weather, and live streams that are stale within minutes; never cached). Each query is labeled at collection time and classified at inference by a lightweight text classifier. Freshness class is the primary input to the risk model and determines which half-life governs staleness decay.

\subsection{Staleness Risk Model}
\label{sec:riskmodel}

\subsubsection{Rule-Based Exponential Decay}

Content change rates follow an approximately exponential distribution across web domains~\cite{cho2003frequency, fetterly2003evolution}. For each class $f$ we fit a half-life $h_f$ from observed change rates, giving:
\begin{equation}
  P_\text{content}(f, \Delta t) = 1 - \exp\!\left( -\frac{\log 2}{h_f}
  \cdot \Delta t \right)
  \label{eq:decay}
\end{equation}
Answer- and URL-tier content are more abstract than raw page content and tend to go stale faster in aggregate, so we apply a tier multiplier $m_\tau$:
\begin{equation}
  \hat{P}(\text{stale} \mid f, \tau, \Delta t) =
  1 - \bigl(1 - P_\text{content}(f, \Delta t)\bigr)^{m_\tau}
  \label{eq:tier}
\end{equation}
with $m_\text{answer} = 1.5$, $m_\text{url\text{-}list} = 1.2$, $m_\text{content} = 1.0$. Section~\ref{sec:design} briefly discusses the sensitivity of this choice; the full sweep is in Appendix~\ref{app:tiermult}.

Half-lives are fit by maximum likelihood from change counts across snapshot reruns at 1h, 12h, 24h, and 7 days, after filtering spurious changes from volatile CDN-served domains. The fitted values are:
\begin{equation}
h=
\left\{
\begin{array}{l}
h_{\text{TIMELESS}} = 22\text{d},\\
h_{\text{SLOW}} = 16\text{d},\\
h_{\text{MEDIUM}} = 15\text{h},\\
h_{\text{FAST}} = 3\text{h},\\
h_{\text{REAL-TIME}} = 30\text{s}
\end{array}
\right.
\label{eq:halflives}
\end{equation}
The 7-day rerun was needed to stabilize the TIMELESS and SLOW estimates, and it corrects an earlier inversion where FAST appeared less volatile than MEDIUM: the most volatile FAST pages, such as live scoreboards, return HTTP 403 to unauthenticated scrapers at short windows and produce no observable change records at 24h. This observability limitation still affects REAL\_TIME, whose 30 second half-life is kept as a prior (Section~\ref{sec:discussion}).

\subsubsection{Learned MLP Risk Model}

The exponential model ignores domain-level volatility within a class and query-conditioned signals. We train a lightweight MLP on 25 input features spanning freshness class (one-hot), log-normalized cache age, per-domain empirical volatility, a 9-way answer type encoding and a temporal keyword indicator, and enrichment features including entity volatility scores, query-query cosine similarity, and lexical and entity overlap. These are passed through three hidden layers of widths [64, 32, 16] with LayerNorm, ReLU, and dropout 0.3, followed by a sigmoid output giving $\hat{P}(\text{stale})$ at the content tier, with Equation~\ref{eq:tier} applied at inference for the other tiers.

Training uses binary cross-entropy with a positive class weight of 8.14 for the 10.9\% positive rate, on a single A6000 GPU for up to 500 epochs with early stopping on validation Brier score. Temperature scaling~\cite{guo2017calibration} with $T = 1.4672$ improves calibration; the final model reaches a Brier score of 0.098 and ECE of 0.140 on a query-cluster-level 70/30 held-out split (2,086 train, 894 test clusters), which prevents leakage of change labels across semantically related queries.

\subsection{Three-Tier Cache Architecture}
\label{sec:tiers}

\textbf{L1: Answer Cache} ($\varepsilon_{L1} = 0.10$) stores the final answer, the URL hashes used to generate it, and a timestamp, and returns without touching the search API, the web, or the LLM. Matching uses BGE-M3 cosine similarity ($\theta_{L1} = 0.40$) plus a named-entity match check using spaCy: a candidate is rejected if either query contains an entity absent from the other, which guards against high similarity pairs that ask about different entities (for example, iPhone 16 versus iPhone 16 Pro pricing). On 20 adversarial pairs mined from the benchmark, cosine similarity alone would approve all 20 as hits; the entity filter blocks 15 of 20, with the misses coming from unusual phrasings where spaCy extracts no entity span (Appendix~\ref{app:adversarial}).

\textbf{L2: URL-List Cache} ($\varepsilon_{L2} = 0.20$, $\theta_{L2} = 0.35$) stores the ordered URL list from a search call. On a hit, the search API is bypassed but pages are still fetched and the LLM still generates a fresh answer, recovering the search cost while avoiding the risk of a different, lower quality URL set for a paraphrased query. No prior semantic cache separates URL discovery from content fetching into distinct cacheable artifacts; this is most valuable at 24 hours, where many MEDIUM and FAST queries carry risk too high for L1 but whose URL sets have not changed even though some page content has shifted.

\textbf{L3: Content Cache} ($\varepsilon_{L3} = 0.35$) stores fetched content keyed by URL hash. The looser budget reflects that partial freshness is acceptable here: even if one page in a result set changed, cached content from the others can still contribute useful context. When risk falls between $\varepsilon_{L3}$ and an upper threshold, FreshCache issues a conditional GET instead of serving stale content or re-fetching unconditionally.

\subsection{Cache Lookup Algorithm}
\label{sec:algorithm}

Algorithm~\ref{alg:lookup} implements the decision rule in Equation~\ref{eq:decision} with graceful degradation across tiers. Lookups are ordered from most to least aggressive: an L1 hit returns in 5ms; an L2 hit saves the search call while regenerating the answer; an L3 hit saves an individual fetch. In the best case every URL is an L3 hit and total latency is just the LLM call; in the worst case every URL misses and latency matches the NoCache baseline.

\begin{algorithm}[!t]
\caption{FreshCache Lookup}
\label{alg:lookup}
\begin{algorithmic}[1]
\small
\Require Query $q$, freshness class $f$, caches $(L1,L2,L3)$, time $t_0$
\Ensure Answer $A$

\If{$f = \text{REAL\_TIME}$} \Return \textsc{FullPipeline}$(q)$ \EndIf

\State $(q',t) \gets \textsc{NN}(q,L1)$
\If{$\mathrm{sim}(q,q') \ge \theta_{L1}$ \textbf{and} $\hat{P}(\mathrm{stale}\mid f,m_a,t_0{-}t) \le \varepsilon_{L1}$}
  \Return $L1[q'].\mathrm{answer}$ \Comment{L1 hit}
\EndIf

\State $(q'',t') \gets \textsc{NN}(q,L2)$
\If{$\mathrm{sim}(q,q'') \ge \theta_{L2}$ \textbf{and} $\hat{P}(\mathrm{stale}\mid f,m_u,t_0{-}t') \le \varepsilon_{L2}$}
  \State $U \gets L2[q''].\mathrm{urls}$ \Comment{L2 hit}
\Else
  \State $U \gets \textsc{SearchAPI}(q)$; \enspace $L2[q] \gets (U,t_0)$
\EndIf

\State $C \gets \emptyset$
\For{each $u \in U$}
  \If{$u\in L3$ \textbf{and} $\hat{P}(\mathrm{stale}\mid f,m_c,t_0{-}L3[u].t) \le \varepsilon_{L3}$}
    \State $C \gets C \cup \{L3[u].\mathrm{content}\}$ \Comment{L3 hit}
  \ElsIf{$u\in L3$ \textbf{and} $\neg\,\operatorname{ConditionalGet}(u, L3[u].\mathrm{etag})$}
    \State $C \gets C \cup \{L3[u].\mathrm{content}\}$ \Comment{Validated reuse}
  \Else
    \State $L3[u] \gets \textsc{Fetch}(u)$; \enspace $C \gets C \cup \{L3[u].\mathrm{content}\}$
  \EndIf
\EndFor

\State $A \gets \textsc{LLM}(q,C)$; \enspace $L1[q] \gets (A,U,t_0)$; \enspace \Return $A$
\end{algorithmic}
\end{algorithm}

\subsection{Lightweight Staleness Validation}
\label{sec:cget}

When a cached URL's risk exceeds $\varepsilon_{L3}$ but a full re-fetch would be wasteful, FreshCache issues a conditional GET using the stored ETag or Last-Modified header, converting a probabilistic estimate into a binary ground truth signal at the cost of one round trip. We model this at 150ms in simulation, but our live-web microbenchmark (Appendix~\ref{app:liveweb}) finds real conditional GET requests average 565ms and only 11.3\% of servers actually return 304. This means content-level validation is a weaker mechanism in production than the simulation assumes, which is precisely why the L1 and L2 tiers, which avoid search and fetch entirely, carry most of FreshCache's value rather than depending on validation.

\subsection{Encoder}
\label{sec:encoder}

All similarity lookups use cosine similarity between BAAI/bge-m3~\cite{chen2024bgem3} embeddings, chosen for multilingual support across our English and Korean queries and for stronger paraphrase detection than monolingual alternatives. Query embeddings are pre-computed into a $31{,}201 \times 31{,}201$ similarity matrix, so lookup during simulation is an O(1) array read; a production deployment would embed incoming queries on the fly with approximate nearest-neighbor search.

\subsection{Design Choices}
\label{sec:design}

We use three tiers rather than one because a single answer-only cache discards search and fetch results that could be partially reused; at 24 hours, FreshCache routes over 20,000 L2 hits for queries whose L1 risk is too high, savings a single-tier system would forfeit. We use separate per-tier error budgets because the answer tier is the most user-visible artifact and deserves the tightest guarantee, while content is an intermediate artifact re-processed by the LLM and tolerates more staleness. We start from a calibrated exponential decay model rather than a learned model because its parameters are interpretable and easy to audit, with the MLP layered on top as an enhancement rather than a replacement; the rule-based model remains available as a fallback when a learned checkpoint is unavailable or out of distribution. The tier multipliers in Equation~\ref{eq:tier} are fixed constants; a sweep over $m_\text{answer}$ shows search savings and stale error are stable across a wide range, so results are not an artifact of this specific choice (Appendix~\ref{app:tiermult}).

\section{Experimental Setup}
\label{sec:setup}

\subsection{FreshCache-Bench}
\label{sec:dataset}

We build FreshCache-Bench from SealQA~\cite{pham2025sealqa}, FreshQA~\cite{vu2024freshllms}, and TemporalAlignQA~\cite{zhao2024setclock} for English queries, plus templated Korean queries generated from entity lists. The benchmark contains 8,072 base queries across five freshness classes, 6,012 English and 246 Korean, each expanded to up to four paraphrases for 31,201 total queries (Table~\ref{tab:dataset}). For each base query we issue a Serper search call and fetch the top URLs at a baseline run and four reruns at approximately 1h, 12h, 24h, and 7 days, hashing each page's main content block after stripping boilerplate; a URL is marked changed if its hash differs from the baseline. After filtering domain-level noise, the benchmark contains 2,995 clean change records across 477 stale URLs at 1 hour, rising to 699 at 24 hours and 1,195 at 7 days, from which the half-lives in Equation~\ref{eq:halflives} are fit.

\begin{table}[!t]
\centering
\caption{FreshCache-Bench dataset composition. URL Cov. = fraction of queries with at least one fetched URL; +Para. = total including paraphrases. Half-lives are calibrated from observed change rates via MLE using snapshot reruns at 1h, 12h, 24h, and 7 days.}
\label{tab:dataset}
\small
\renewcommand{\arraystretch}{1.12}
\setlength{\tabcolsep}{4pt}
\begin{tabular*}{\columnwidth}{@{\extracolsep{\fill}}lrrrr@{}}
\toprule
\textbf{Class} & \textbf{Base} & \textbf{URL Cov.} & \textbf{+Para.} & \textbf{Half-life} \\
\midrule
TIMELESS   & 2{,}000 & 84.6\% & 6{,}929  & 22d \\
SLOW       & 2{,}000 & 83.1\% & 8{,}137  & 16d \\
MEDIUM     & 2{,}000 & 80.5\% & 7{,}626  & 15h \\
FAST       & 2{,}000 & 86.1\% & 8{,}155  & 3h  \\
REAL\_TIME & 72      & 98.3\% & 354      & 30s \\
\midrule
\textbf{Total} & \textbf{8{,}072} & \textbf{84.1\%} & \textbf{31{,}201} & --- \\
\bottomrule
\end{tabular*}
\end{table}

\subsection{Baselines}
\label{sec:baselines}

We compare against NoCache and ExactTTL as lower and upper bounds on staleness and cost; SemanticTTL~\cite{bang2023gptcache}, a GPTCache-style cosine match with per-class TTL; DomainTTL, a per-class TTL scaled by mean domain volatility; TemporalKeywordTTL, which halves the TTL when the query contains a temporal keyword; L3Only and L3+ConditionalGet, which always search but cache content with and without validation; vCache~\cite{schroeder2025vcache} with $\varepsilon = 0.10$; SCALM~\cite{li2024scalm}; TieredFixedTTL, which shares FreshCache's tier structure but replaces the risk model with a binary TTL gate at the same half-lives, isolating the risk model's contribution; and OracleFreshness, a non-deployable upper bound using ground-truth per-URL labels. We also report FreshCache\_NoCalib, using naive prior half-lives instead of MLE-fit values, to isolate the calibration contribution.

\subsection{Implementation and Metrics}
\label{sec:impl}

All simulations run on a single CPU node over the pre-computed similarity matrix; embeddings are generated once with BAAI/bge-m3 across two A6000 GPUs. The MLP is trained on a single A6000 with AdamW at learning rate $10^{-3}$. Key hyperparameters are listed in Appendix~\ref{app:impl}.

We report three metrics. Stale error rate is the fraction of cache hits whose underlying URL content changed since population, using SHA-256 content hash comparison as a reproducible, conservative proxy for answer-affecting staleness; Section~\ref{sec:llmjudge} validates this proxy against an LLM judge. Search savings is the fraction of queries for which a fresh search call was avoided. Base query p50 latency saving is the reduction in median end-to-end latency on base queries only, reported separately because the overall p50 is dominated by paraphrase L1 hits at 5ms and would overstate production benefit by roughly a factor of four. All experiments run at $t=1$h and $t=24$h; 24 hours is our primary window and design target.

\subsection{Metric Validation}
\label{sec:llmjudge}

Hash comparison counts any content change as staleness, which conflates genuine information updates with cosmetic changes such as rotating banners or view counters. To quantify this, we ran an LLM judge (Llama-3.2-3B-Instruct, local) on all 396 fully evaluable URL-query pairs with a hash-detected change at 24 hours, asking whether the change would make a cached answer factually wrong. Only 34.3\% of hash-detected changes are answer-affecting (Appendix~\ref{app:llmjudge} for the per-class breakdown), which places FreshCache\_MLP's true answer-level stale error at approximately 0.034\%, a factor of 2.9 below the reported hash-based figure. A separate end-to-end grading experiment on 250 stratified queries, comparing answers generated from baseline versus 24-hour rerun evidence, found 95.6\% overall cache-hit answer accuracy and only 17.2\% of hash-detected changes resulting in an actual answer error, corroborating the hash metric as a conservative upper bound. All comparison tables in this paper use the hash-based proxy for reproducibility across methods.

\subsection{Live-Web and External Validation}
\label{sec:liveweb_cise}

The 800ms fetch and 150ms conditional GET latencies used in simulation are modelling constants. A live-web microbenchmark on 149 URLs finds median cold-fetch latency of 915ms, close to the simulated value, but p95 latency of 2,771ms, 3.5 times the constant, meaning the simulation understates FreshCache's advantage at high percentiles (Appendix~\ref{app:liveweb}). We also evaluate cache-induced stale error directly on two external QA benchmarks, EvolvingQA and DailyQA, neither used for calibration. FreshCache\_MLP's hit rate drops sharply on both, since its learned features are tuned to FreshCache-Bench's domain volatility distribution rather than real-world event timing; the rule-based model degrades more gracefully and remains the recommended fallback outside the training distribution (Appendix~\ref{app:cise}).

\section{Results and Discussion}
\label{sec:results}

\subsection{Main Results}
\label{sec:mainresults}

\begin{figure}[!t]
    \centering
    \includegraphics[width=\columnwidth]{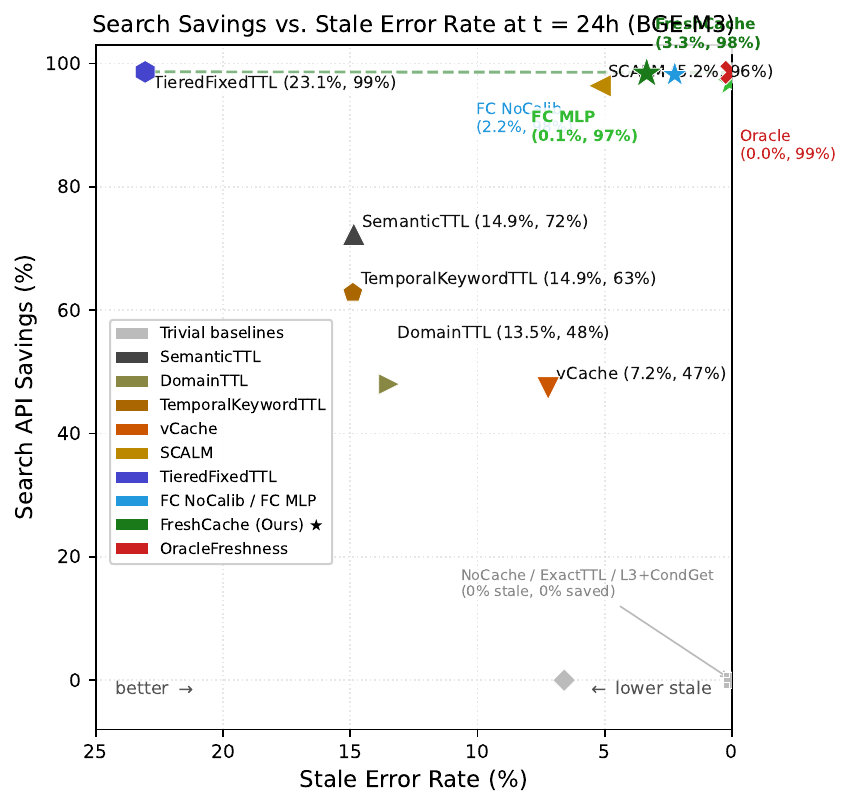}
    \caption{Search API savings versus stale error rate at $t = 24$h using BGE-M3. FreshCache\_MLP achieves the best deployable freshness--efficiency trade-off, with 0.1\% stale error and 97\% search savings. OracleFreshness represents the non-deployable upper bound.}
    \label{fig:pareto}
\end{figure}

Table~\ref{tab:main24} reports results at $t=24$h with BGE-M3. FreshCache\_MLP saves 97\% of search calls at 0.1\% stale error and is Pareto-dominant over every baseline (Figure~\ref{fig:pareto}): SemanticTTL reaches 14.9\% stale at 72\% saved, vCache 7.2\% at 47\%, and SCALM 5.2\% at 96\%. These savings hold on base queries alone: excluding all 24,943 paraphrases gives 97.3\% saved at 0.0\% stale error, so the headline result is not an artifact of paraphrases arriving at near-zero cache age.

To confirm statistical reliability given the finite evaluation set, we
ran a nonparametric bootstrap (30 resamples, query records resampled
with replacement, stale URL set fixed at 727 URLs, seed=42).
FreshCache\_MLP achieves a bootstrap 95\% confidence interval of
[0.08\%, 0.16\%] on overall stale error and [0.05\%, 0.20\%] on the
FAST class, the most volatile and most sensitive to workload variation.
The rule-based FreshCache shows a 95\% CI of [2.68\%, 5.34\%] overall.
The gap between the two methods is statistically robust: no bootstrap
resample brought FreshCache\_MLP above 0.20\% stale error or
FreshCache\_Full below 2.68\%.

The rule-based half-life in Table~\ref{tab:main24} is fit under a temporal holdout, using rerun\_1h observations only and evaluating on rerun\_24h, which the fit never sees. Under this honest holdout, stale error is 3.3\% at 98.4\% saved, higher than the 1.4\% obtained when the same snapshot is used for both fitting and evaluation, because a single exponential rate cannot jointly capture short-term churn and slower long-term drift~\cite{olston2008recrawl}. FreshCache\_MLP uses a separate query-cluster holdout (Section~\ref{sec:riskmodel}) and is unaffected by this issue, remaining our primary deployable result. A TTL sweep across four baselines and eight scale factors confirms no configuration simultaneously beats FreshCache\_Full on both stale error and search savings (Appendix~\ref{app:ttlsweep}).

\begin{table}[!t]
\centering
\caption{Results at $t = 24$h on all classes using BGE-M3 encoder
(31,201 queries, 727 stale URLs). S/q = search calls per query;
Stale = stale error rate; Saved = search savings. The primary deployable FreshCache result is shown in \textbf{bold}.
$\dagger$ = oracle upper bound, not deployable.}
\label{tab:main24}
\small
\renewcommand{\arraystretch}{1.12}
\setlength{\tabcolsep}{4pt}
\begin{tabular*}{\columnwidth}{@{\extracolsep{\fill}}lccc@{}}
\toprule
\textbf{Method} & \textbf{S/q} & \textbf{Stale} & \textbf{Saved} \\
\midrule
SemanticTTL      & 0.28 & 14.9\% & 72\% \\
DomainTTL        & 0.52 & 13.5\% & 48\% \\
TempKeyTTL       & 0.37 & 14.9\% & 63\% \\
vCache           & 0.53 & 7.2\%  & 47\% \\
SCALM            & 0.04 & 5.2\%  & 96\% \\
TieredTTL        & 0.01 & 23.1\% & 99\% \\
\midrule
FC\_NoCalib      & 0.02 & 2.2\%  & 98\% \\
FreshCache       & 0.02 & 3.3\%  & 98\% \\
\textbf{FC\_MLP} & \textbf{0.03} & \textbf{0.1\%} & \textbf{97\%} \\
\midrule
Oracle$^\dagger$ & 0.01 & \textit{0.0\%} & \textit{99\%} \\
\bottomrule
\end{tabular*}

\vspace{1mm}
\begin{minipage}{\columnwidth}
\scriptsize
TempKeyTTL = TemporalKeywordTTL; TieredTTL = TieredFixedTTL;
FC\_NoCalib = FreshCache\_NoCalib; FC\_MLP = FreshCache\_MLP. Results at $t=1$h follow the same ordering and appear in Appendix~\ref{app:t1h}.
\end{minipage}
\end{table}

\subsection{Per-Class Breakdown}
\label{sec:perclass}

Table~\ref{tab:perclass} breaks stale error down by class. TIMELESS and SLOW content rarely changes, and all methods stay under 9\% except SemanticTTL and SCALM, which cannot separate stable from volatile entries within a class. MEDIUM is where temporal risk estimation matters most: the rule-based model permits some reuse at 4.3\% stale error under the holdout fit, while the MLP recovers 3,918 MEDIUM L1 hits at 0.0\% stale error by conditioning on answer type and entity volatility, though Appendix~\ref{app:cise} shows this within-class discrimination does not transfer to out-of-distribution query sets. FAST shows the largest holdout penalty, rising to 5.4\% stale error, since it has the shortest half-life and the least observation history at the 1-hour fit window. REAL\_TIME is never cached by design and reports 0\% stale error and 0\% savings for every method.

\begin{table*}[!t]
\centering
\caption{Stale error rate per freshness class at $t = 24$h using BGE-M3. Search savings for FreshCache\_MLP are shown for reference. $\ast$ = SemanticTTL TTL for FAST expired at 24h, making zero caching decisions. $\ddagger$ = oracle upper bound, not deployable. FreshCache values use half-lives fit under the temporal holdout described in Section~\ref{sec:mainresults}.}
\label{tab:perclass}
\scriptsize
\renewcommand{\arraystretch}{1.12}
\setlength{\tabcolsep}{4pt}
\begin{tabular*}{\textwidth}{@{\extracolsep{\fill}}lrrrrrr@{}}
\toprule
\textbf{Method} 
& \textbf{TIMELESS} 
& \textbf{SLOW} 
& \textbf{MEDIUM}
& \textbf{FAST} 
& \textbf{ALL} 
& \textbf{Search Saved} \\
\midrule
SemanticTTL  & 2.6\% & 9.5\% & 5.9\% & 0.0\%$^\ast$ & 14.9\% & --- \\
vCache       & 2.3\% & 6.2\% & 8.4\% & 10.8\%        & 7.2\%  & --- \\
SCALM        & 1.6\% & 8.9\% & 5.8\% & 3.9\%         & 5.2\%  & --- \\
\midrule
FreshCache   & 1.5\% & 2.9\% & 4.3\% & 5.4\%         & 3.3\%  & 95--99\% \\
\textbf{FC\_MLP} 
             & \textbf{0.0\%} 
             & \textbf{0.2\%} 
             & \textbf{0.0\%} 
             & \textbf{0.1\%} 
             & \textbf{0.1\%} 
             & \textbf{95--97\%} \\
\midrule
Oracle$^\ddagger$ 
             & \textit{0.0\%} 
             & \textit{0.0\%} 
             & \textit{0.0\%} 
             & \textit{0.0\%} 
             & \textit{0.0\%} 
             & \textit{99\%} \\
\bottomrule
\end{tabular*}

\vspace{1mm}
\begin{minipage}{\textwidth}
\scriptsize
FC\_MLP = FreshCache\_MLP. Search Saved denotes search API savings. All methods use the BGE-M3 encoder.
\end{minipage}
\end{table*}

\subsection{Tier Activation}
\label{sec:tiershift}

Table~\ref{tab:tiers} shows per-class tier activation at both windows. At $t=1$h, TIMELESS, SLOW, and MEDIUM queries are served from L1, while FAST already shows zero L1 hits because its 3h half-life pushes risk above budget even at short ages. By $t=24$h, MEDIUM joins FAST with zero L1 hits, but 7,582 MEDIUM queries route through L2, bypassing search while still regenerating the answer from fresh content; FAST falls through to a full search at both windows. This tier shift is not available to single-tier answer caches, which must either approve reuse or fall back entirely. The behavior is grounded in a workload characterization showing that TIMELESS and SLOW query pairs with high semantic similarity reliably share URL sets, while MEDIUM and FAST pairs do not, which is why L2 reuse is selective rather than uniform (Appendix~\ref{app:workload}).

\begin{table*}[!t]
\centering
\caption{FreshCache (rule-based) tier hits per freshness class using BGE-M3 on 31,201 queries. L1 = answer hits, L2 = URL-list hits, and L3 = content hits. Per-class rows are isolated single-class runs, each starting from an empty cache. $\dagger$ marks the Total row, which is measured from a single joint run over all 31,201 queries sharing one cache; because an L1 entry populated while serving one class can still be reused when a later query from another class arrives, this Total is not the arithmetic sum of the isolated per-class rows above it.}
\label{tab:tiers}
\scriptsize
\renewcommand{\arraystretch}{1.15}
\setlength{\tabcolsep}{8pt}
\begin{tabular*}{\textwidth}{@{\extracolsep{\fill}}lrrrrrr@{}}
\toprule
\multirow{2}{*}{\textbf{Class}} 
& \multicolumn{3}{c}{\textbf{$t = 1$h}} 
& \multicolumn{3}{c}{\textbf{$t = 24$h}} \\
\cmidrule(lr){2-4} \cmidrule(lr){5-7}
& \textbf{L1} & \textbf{L2} & \textbf{L3}
& \textbf{L1} & \textbf{L2} & \textbf{L3} \\
\midrule
TIMELESS   & 1{,}691 & 5{,}111  & 5{,}769  & 1{,}691            & 5{,}111  & 5{,}769  \\
SLOW       & 3{,}238 & 4{,}838  & 5{,}205  & 3{,}238 & 4{,}838  & 5{,}205  \\

MEDIUM     & 4{,}090 & 3{,}494  & 3{,}650  & 0                  & 7{,}582  & 9{,}036  \\
FAST       & 0       & 0        & 10{,}206 & 0                  & 0        & 9{,}021  \\
REAL\_TIME & 0       & 0        & 0        & 0                  & 0        & 0        \\
\midrule
\textbf{Total$^\dagger$} 
           & 5{,}072 & 25{,}562 & 29{,}455 
           & 10{,}182 & 20{,}400 & 21{,}453 \\
\bottomrule
\end{tabular*}
\end{table*}

\subsection{Discussion}
\label{sec:discussion}

SCALM's frequency-weighted selection tends to serve the most popular cached entry above a similarity threshold, and popular entries skew toward TIMELESS and SLOW clusters; when a MEDIUM query arrives with modest similarity to a TIMELESS entry, SCALM serves a semantically plausible but temporally stale answer. FreshCache avoids this by conditioning reuse on class-specific staleness probability rather than hit frequency. At $t=1$h, the rule-based model shows only a small advantage over SemanticTTL on MEDIUM queries (4.1\% versus 4.7\% stale), because the 15-hour half-life implies low risk at short ages; the MLP resolves this using query-conditioned features, and at $t=24$h the class-level rule already blocks all MEDIUM L1 reuse once the half-life has expired. The reported p50 latency saving of roughly 99.8\% is inflated by paraphrase L1 hits at 5ms; a more honest production estimate, applying the measured fetch constants from Appendix~\ref{app:liveweb} to FreshCache\_MLP's actual hit and miss ratios, gives approximately 80\% median latency reduction, with true high-percentile savings larger than this simulation reports since real p95 fetch latency is 3.5 times the simulated constant. Finally, since only 34.3\% of hash-flagged staleness is answer-affecting (Section~\ref{sec:llmjudge}), the absolute stale error numbers reported throughout this paper should be read as conservative upper bounds rather than exact production error rates, and the correction factor applies equally across methods so the relative ordering in every table is unaffected.

\section{Ablation Study}
\label{sec:ablation}

\subsection{Encoder Contribution}
\label{sec:ablation_encoder}

A concern with comparing systems on a stronger encoder is that encoder quality alone could explain better performance. Table~\ref{tab:ablation_encoder} compares MiniLM-L6-v2 against BGE-M3 with all other components fixed, both calibrated under the same temporal holdout. Upgrading the encoder \emph{hurts} SemanticTTL, raising its stale error from 10.4\% to 14.9\%, since a stronger encoder finds more similar pairs but a fixed TTL cannot filter the additional hits by risk. vCache is essentially flat across encoders, 7.3\% to 7.2\%. FreshCache shows the opposite pattern entirely: stale error improves from 3.8\% to 3.3\% while search savings rise from 96\% to 98\%, and the MLP variant reaches 0.1\% stale error at 97\% saved regardless of encoder. This asymmetry is the key finding: a stronger encoder without temporal gating makes staleness meaningfully worse, while the same encoder with temporal gating holds staleness steady or improves it while adding search savings, confirming that FreshCache's architecture, not encoder quality, drives its advantage over SemanticTTL. BGE-M3 is also necessary for our multilingual setting: MiniLM effectively prevents cache hits on Korean queries.

\begin{table}[!t]
\centering
\caption{Encoder effect on stale error at $t = 24$h on all classes using rule-based FreshCache unless noted. $\uparrow$ means the encoder upgrade helps; $\downarrow$ means it hurts.}
\label{tab:ablation_encoder}
\small
\renewcommand{\arraystretch}{1.12}
\setlength{\tabcolsep}{4pt}
\begin{tabular*}{\columnwidth}{@{\extracolsep{\fill}}lcccc@{}}
\toprule
\multirow{2}{*}{\textbf{Method}} 
& \multicolumn{2}{c}{\textbf{MiniLM-L6}} 
& \multicolumn{2}{c}{\textbf{BGE-M3}} \\
\cmidrule(lr){2-3} \cmidrule(lr){4-5}
& \textbf{Stale} & \textbf{Saved} 
& \textbf{Stale} & \textbf{Saved} \\
\midrule
SemanticTTL & 10.4\% & 69\% & 14.9\%$\downarrow$ & 72\% \\
vCache      & 7.3\%  & 49\% & 7.2\%              & 47\% \\
\midrule
\textbf{FreshCache} 
            & 3.8\%  & 96\% 
            & \textbf{3.3\%}$\uparrow$ 
            & \textbf{98\%}$\uparrow$ \\
\bottomrule
\end{tabular*}

\vspace{1mm}
\begin{minipage}{\columnwidth}
\scriptsize
FreshCache\_MLP with BGE-M3 reaches 0.1\% stale error at 97\% saved, not shown as a separate row since it uses BGE-M3 only.
\end{minipage}
\end{table}

\subsection{Tier Contribution}
\label{sec:ablation_tiers}

Table~\ref{tab:ablation_tiers} isolates each tier's contribution. TieredFixedTTL adds the full L2 and L3 tiers to SemanticTTL's structure but keeps a fixed TTL gate, reaching 99\% search savings at 23.1\% stale error, the highest in the table despite having every tier active. This is the central result of this ablation: adding tiers reduces search cost, but only the risk model reduces stale error. L3Only, which always pays the search cost but risk-gates content, achieves 4.2\% stale error at zero search savings, confirming the L2 tier is what lets FreshCache bypass the search API without sacrificing freshness; it routes roughly 20,400 queries through L2 rather than L1 at 24 hours, producing answers current as of query time from re-fetched content. L3+ConditionalGet achieves zero stale error by construction but zero search savings, and FreshCache approximates this guarantee while adding 97 to 98\% search savings on top.

\begin{table}[!t]
\centering
\caption{Tier contribution ablation at $t = 24$h on all classes using BGE-M3. S/q = search calls per query; Stale = stale error rate; Saved = search savings. $\star$ = zero stale errors by construction.}
\label{tab:ablation_tiers}
\small
\renewcommand{\arraystretch}{1.12}
\setlength{\tabcolsep}{3pt}
\begin{tabular*}{\columnwidth}{@{\extracolsep{\fill}}llccc@{}}
\toprule
\textbf{Method} & \textbf{Tiers} & \textbf{S/q} & \textbf{Stale} & \textbf{Saved} \\
\midrule
NoCache        & None        & 1.00 & 0.0\%         & 0\%  \\
SemanticTTL    & L1+TTL      & 0.28 & 14.9\%        & 72\% \\
TieredTTL      & L1--L3+TTL  & 0.01 & 23.1\%        & 99\% \\
L3Only         & L3          & 1.00 & 4.2\%         & 0\%  \\
L3+CondGet     & L3+valid.   & 1.00 & 0.0\%$^\star$ & 0\%  \\
\midrule
FreshCache     & L1--L3+risk & 0.02 & 3.3\%         & 98\% \\
\textbf{FC\_MLP} 
               & \textbf{L1--L3+MLP} 
               & \textbf{0.03} & \textbf{0.1\%} & \textbf{97\%} \\
\bottomrule
\end{tabular*}

\vspace{1mm}
\begin{minipage}{\columnwidth}
\scriptsize
TieredTTL = TieredFixedTTL; valid. = validation; 
FC\_MLP = FreshCache\_MLP. L1--L3 means all three cache tiers are active.
\end{minipage}
\end{table}

\subsection{Temporal Risk Gate}
\label{sec:ablation_riskgate}

SemanticTTL is effectively FreshCache with the risk gate removed and replaced by class-level TTLs, so comparing them isolates the gate's effect. At $t=1$h the gate provides little benefit, since fitted half-lives already predict low staleness at short ages and the risk model approves nearly the same L1 hits as SemanticTTL. At $t=24$h the gate becomes decisive: overall stale error falls from 14.9\% to 3.3\%, an 11.6 point reduction, driven mainly by MEDIUM falling to 0.0\% as the model blocks L1 reuse once the 15h half-life has expired, and SLOW falling to 2.9\% as content is routed through L2 and L3. TIMELESS keeps near-perfect search savings throughout, since its risk stays below the 10\% budget at both windows. The full per-class comparison at both time windows is in Appendix~\ref{app:riskgate}.

\FloatBarrier
\subsection{Learned MLP}
\label{sec:ablation_mlp}

\begin{table*}
\centering
\caption{Rule-based vs. MLP risk model per freshness class at $t = 24$h using BGE-M3. $\Delta$L1 and $\Delta$Stale show the change when switching from the rule-based model to the 25-dimensional MLP.}
\label{tab:ablation_mlp}
\scriptsize
\renewcommand{\arraystretch}{1.12}
\setlength{\tabcolsep}{4pt}
\begin{tabular*}{\textwidth}{@{\extracolsep{\fill}}lrrrrrr@{}}
\toprule
\multirow{2}{*}{\textbf{Class}} 
& \multicolumn{2}{c}{\textbf{Rule-based}} 
& \multicolumn{2}{c}{\textbf{MLP}} 
& \multicolumn{2}{c}{\textbf{Change}} \\
\cmidrule(lr){2-3} \cmidrule(lr){4-5} \cmidrule(lr){6-7}
& \textbf{L1} & \textbf{Stale} 
& \textbf{L1} & \textbf{Stale} 
& \textbf{$\Delta$L1} & \textbf{$\Delta$Stale} \\
\midrule
TIMELESS & 1{,}691 & 1.5\% & 1{,}467 & 0.0\% & $-224$     & $-1.5$ \\
SLOW     & 3{,}238 & 2.9\% & 828     & 0.2\% & $-2{,}410$ & $-2.7$ \\
MEDIUM   & 0       & 4.3\% & 3{,}918 & 0.0\% & $+3{,}918$ & $-4.3$ \\
FAST     & 0       & 5.4\% & 956     & 0.1\% & $+956$     & $-5.3$ \\
\midrule
\textbf{ALL} 
         & 10{,}182 & \textbf{3.3\%} 
         & 4{,}131  & \textbf{0.1\%} 
         & $-6{,}051$ & $-3.2$ \\
\bottomrule
\end{tabular*}
\end{table*}

The rule-based model gives identical risk to any two queries in the same class at the same age; the MLP's answer type, entity volatility, and similarity features close this gap. Table~\ref{tab:ablation_mlp} compares the two at $t=24$h. The MLP approves 828 additional SLOW L1 hits (0.2\% stale) on top of the 3,238 the rule-based model already allows, and opens L1 entirely for MEDIUM (3,918 hits, 0.0\% stale) and FAST (956 hits, 0.1\% stale), where the rule-based model routes both classes through L2 and L3 because class-level risk exceeds budget. TIMELESS is the one class where the MLP is more conservative, trading a small drop in L1 hit count for zero stale error. Overall the MLP reduces stale error from 3.3\% to 0.1\% at $t=24$h, and from 2.6\% to 2.0\% at $t=1$h, at a cost of roughly one point of search savings.

\subsection{Summary}
\label{sec:ablation_summary}

The four ablations establish a clear hierarchy: the temporal risk gate reduces stale error by 11.6 points over SemanticTTL; the L2 tier is what opens a search-saving path for SLOW, MEDIUM, and FAST queries once direct answer reuse becomes too risky; and the learned MLP reduces stale error a further 3.2 points through within-class discrimination the rule-based model cannot express.

\section{Conclusion}
\label{sec:conclusion}

We argued that caching for open-web RAG is incomplete when framed only as similarity-based retrieval, since the right question before serving a cached result is whether the underlying evidence is still likely to be accurate. FreshCache operationalizes this as a risk-constrained temporal reuse decision across three tiers, each with its own staleness budget, backed by a calibrated decay model and an optional learned MLP. On FreshCache-Bench, FreshCache\_MLP reaches 0.1\% stale error at 97\% search savings at the 24-hour window, within 2 points of the oracle ceiling, and an LLM-judge correction places true answer-level stale error near 0.034\%. Four ablations show this result depends jointly on the tier structure and the risk gate, neither of which is sufficient alone.

Three limitations remain. The rule-based model offers limited improvement over SemanticTTL at $t=1$h, which the MLP substantially addresses; our simulated p50 latency saving is inflated by paraphrase hits, with an honest production estimate closer to 80\% (Section~\ref{sec:discussion}); and conditional GET, while part of the content tier design, proved largely ineffective on the live web, reinforcing that the L1 and L2 tiers should carry the majority of production savings rather than validation (Appendix~\ref{app:liveweb}). Future work includes broader domain coverage for FreshCache-Bench, integration with authenticated data feeds to address the FAST observability bias, adaptive error budgets set from observed stale rates rather than fixed at deployment, and richer entity-level features to help the learned model generalize beyond its training distribution, as motivated by its limited transfer in Appendix~\ref{app:cise}.


\bibliography{custom}

@inproceedings{lewis2020rag,
  author    = {Lewis, Patrick and Perez, Ethan and Piktus, Aleksandra and
               Petroni, Fabio and Karpukhin, Vladimir and Goyal, Naman and
               K{\"u}ttler, Heinrich and Lewis, Mike and Yih, Wen-tau and
               Rockt{\"a}schel, Tim and Riedel, Sebastian and Kiela, Douwe},
  title     = {Retrieval-Augmented Generation for Knowledge-Intensive {NLP} Tasks},
  booktitle = {Advances in Neural Information Processing Systems},
  volume    = {33},
  pages     = {9459--9474},
  year      = {2020},
  publisher = {Curran Associates, Inc.}
}

@preprint{gao2023survey,
  author        = {Gao, Yunfan and Xiong, Yun and Gao, Xinyu and Jia, Kangxiang and
                   Pan, Jinliu and Bi, Yuxi and Dai, Yi and Sun, Jiawei and Wang, Haofen},
  title         = {Retrieval-Augmented Generation for Large Language Models: A Survey},
  year          = {2023},
  archivePrefix = {arXiv},
  eprint        = {2312.10997}
}

@preprint{nakano2021webgpt,
  author        = {Nakano, Reiichiro and Hilton, Jacob and Balwit, Avital and Wu,
                   Jeff and Ouyang, Long and Kim, Christina and Hesse, Christopher and
                   Jain, Shantanu and Kosaraju, Vineet and Saunders, William and
                   others},
  title         = {{WebGPT}: Browser-Assisted Question-Answering with Human Feedback},
  year          = {2021},
  archivePrefix = {arXiv},
  eprint        = {2112.09332}
}

@preprint{lazaridou2022internet,
  author        = {Lazaridou, Angeliki and Gribovskaya, Elena and Stokowiec,
                   Wojciech and Grigorev, Nikolai},
  title         = {Internet-Augmented Language Models through Few-Shot Prompting
                   for Open-Domain Question Answering},
  year          = {2022},
  archivePrefix = {arXiv},
  eprint        = {2203.05115}
}

@inproceedings{mallen2023trust,
  author    = {Mallen, Alex and Asai, Akari and Zhong, Victor and Das,
               Rajarshi and Khashabi, Daniel and Hajishirzi, Hannaneh},
  title     = {When Not to Trust Language Models: Investigating Effectiveness
               of Parametric and Non-Parametric Memories},
  booktitle = {Proceedings of the 61st Annual Meeting of the Association for
               Computational Linguistics},
  pages     = {9053--9070},
  year      = {2023}
}

@inproceedings{bang2023gptcache,
  author    = {Bang, Fu},
  title     = {{GPTCache}: An Open-Source Semantic Cache for {LLM} Applications
               Enabling Faster Answers and Cost Savings},
  booktitle = {Proceedings of the 3rd Workshop for Natural Language Processing
               Open Source Software ({NLP-OSS} 2023)},
  pages     = {212--218},
  year      = {2023},
  address   = {Singapore},
  publisher = {Association for Computational Linguistics}
}

@preprint{zhu2024semsimilarity,
  author        = {Zhu, Hanlin and Zhu, Banghua},
  title         = {Efficient Prompt Caching via Embedding Similarity},
  year          = {2024},
  archivePrefix = {arXiv},
  eprint        = {2402.01173}
}

@article{li2024scalm,
  author  = {Jiaxing Li and Chi Xu and Feng Wang and Isaac M. von Riedemann and Cong Zhang and Jiangchuan Liu},
  title   = {{SCALM}: Towards Semantic Caching for Automated Chat Services with Large Language Models},
  journal = {arXiv preprint arXiv:2406.00025},
  year    = {2024}
}

@preprint{schroeder2025vcache,
  author        = {Schroeder, Luis Gaspar and Desai, Aditya and Cuadron, Alejandro
                   and Chu, Kyle and Liu, Shu and Zhao, Mark and Krusche, Stephan and
                   Kemper, Alfons and Stoica, Ion and Zaharia, Matei and
                   Gonzalez, Joseph E.},
  title         = {{vCache}: Verified Semantic Prompt Caching},
  year          = {2025},
  archivePrefix = {arXiv},
  eprint        = {2502.03771}
}

@preprint{chan2025cag,
  author        = {Chan, Brian J. and others},
  title         = {Don't Do {RAG}: When Cache-Augmented Generation is All You Need
                   for Knowledge Tasks},
  year          = {2025},
  archivePrefix = {arXiv},
  eprint        = {2412.15605}
}

@preprint{jin2024ragcache,
  author        = {Jin, Chao and others},
  title         = {{RAGCache}: Efficient Knowledge Caching for Retrieval-Augmented
                   Generation},
  year          = {2024},
  archivePrefix = {arXiv},
  eprint        = {2404.12457}
}

@inproceedings{yao2025cacheblend,
  author    = {Yao, Jiayi and Li, Hanchen and Liu, Yuhan and Ray, Siddhant and
               Cheng, Yihua and Zhang, Qizheng and Du, Kuntai and Lu, Shan and
               Ananthanarayanan, Ganesh},
  title     = {{CacheBlend}: Fast Large Language Model Serving for {RAG} with
               Cached Knowledge Fusion},
  booktitle = {Proceedings of the Twentieth European Conference on Computer
               Systems},
  pages     = {94--109},
  year      = {2025}
}

@inproceedings{vu2024freshllms,
  author    = {Vu, Tu and Iyyer, Mohit and Wang, Xuezhi and Constant, Nick and
               Wei, Jerry and Wei, Jason and Tar, Chris and Sung, Yun-Hsuan and
               Zhou, Denny and Le, Quoc V. and Luong, Thang},
  title     = {{FreshLLMs}: Refreshing Large Language Model Knowledge through
               Search Engine Augmentation},
  booktitle = {Findings of the Association for Computational Linguistics: EMNLP
               2023},
  pages     = {13697--13720},
  year      = {2024}
}

@inproceedings{reimers2019sbert,
  author    = {Reimers, Nils and Gurevych, Iryna},
  title     = {Sentence-{BERT}: Sentence Embeddings using Siamese {BERT}-Networks},
  booktitle = {Proceedings of the 2019 Conference on Empirical Methods in Natural
               Language Processing},
  pages     = {3982--3992},
  year      = {2019}
}

@preprint{chen2024bgem3,
  author        = {Chen, Jianlv and Xiao, Shitao and Zhang, Peitian and Luo, Kun and
                   Lian, Defu and Liu, Zheng},
  title         = {{BGE} {M3}-Embedding: Multi-Lingual, Multi-Functionality,
                   Multi-Granularity Text Embeddings Through Self-Knowledge Distillation},
  year          = {2024},
  archivePrefix = {arXiv},
  eprint        = {2309.07597}
}

@article{cho2003frequency,
  author  = {Cho, Junghoo and Garcia-Molina, Hector},
  title   = {Estimating Frequency of Change},
  journal = {ACM Transactions on Internet Technology},
  volume  = {3},
  number  = {3},
  pages   = {256--290},
  year    = {2003}
}

@inproceedings{karpukhin2020dpr,
  author    = {Karpukhin, Vladimir and O\u{g}uz, Barlas and Min, Sewon and
               Lewis, Patrick and Wu, Ledell and Edunov, Sergey and Chen,
               Danqi and Yih, Wen-tau},
  title     = {Dense Passage Retrieval for Open-Domain Question Answering},
  booktitle = {Proceedings of the 2020 Conference on Empirical Methods in
               Natural Language Processing},
  pages     = {6769--6781},
  year      = {2020},
  address   = {Online},
  publisher = {Association for Computational Linguistics}
}

@article{dhingra2022time,
  author    = {Dhingra, Bhuwan and Cole, Jeremy R. and Eisenschlos, Julian
               Martin and Gillick, Daniel and Eisenstein, Jacob and
               Cohen, William W.},
  title     = {Time-Aware Language Models as Temporal Knowledge Bases},
  journal   = {Transactions of the Association for Computational Linguistics},
  volume    = {10},
  pages     = {257--273},
  year      = {2022},
  publisher = {MIT Press},
  doi       = {10.1162/tacl_a_00459}
}

@inproceedings{meng2022rome,
  author    = {Meng, Kevin and Bau, David and Andonian, Alex and
               Belinkov, Yonatan},
  title     = {Locating and Editing Factual Associations in {GPT}},
  booktitle = {Advances in Neural Information Processing Systems},
  volume    = {35},
  pages     = {17359--17372},
  year      = {2022},
  publisher = {Curran Associates, Inc.}
}

@inproceedings{tan2023temporal,
  author    = {Tan, Qingyu and Ng, Hwee Tou and Bing, Lidong},
  title     = {Towards Benchmarking and Improving the Temporal Reasoning
               Capability of Large Language Models},
  booktitle = {Proceedings of the 61st Annual Meeting of the Association for
               Computational Linguistics (Volume 1: Long Papers)},
  pages     = {14820--14835},
  year      = {2023},
  address   = {Toronto, Canada},
  publisher = {Association for Computational Linguistics}
}

@inproceedings{fetterly2003evolution,
  author    = {Fetterly, Dennis and Manasse, Mark and Najork, Marc and
               Wiener, Janet},
  title     = {A Large-Scale Study of the Evolution of {Web} Pages},
  booktitle = {Proceedings of the 12th International Conference on World
               Wide Web},
  pages     = {669--678},
  year      = {2003},
  address   = {Budapest, Hungary},
  publisher = {ACM Press}
}

@inproceedings{kasai2023realtime,
  author    = {Kasai, Jungo and Sakaguchi, Keisuke and Takahashi, Yoichi and
               Le Bras, Ronan and Asai, Akari and Yu, Xinyan and Radev,
               Dragomir and Smith, Noah A. and Choi, Yejin and Inui, Kentaro},
  title     = {{RealTime QA}: What's the Answer Right Now?},
  booktitle = {Advances in Neural Information Processing Systems},
  volume    = {36},
  year      = {2023},
  publisher = {Curran Associates, Inc.}
}

@preprint{pham2025sealqa,
  author        = {Pham, Thinh and Nguyen, Phi Tran Anh and
                   Nguyen, Anh Tuan and Luu, Anh Tuan},
  title         = {{SealQA}: Raising the Bar for Reasoning in
                   Search-Augmented Language Models},
  year          = {2025},
  archivePrefix = {arXiv},
  eprint        = {2506.01062}
}

@inproceedings{zhao2024setclock,
  author    = {Zhao, Bowen and Brumbaugh, Zander and
               Wang, Yizhong and Hajishirzi, Hannaneh and
               Smith, Noah A.},
  title     = {Set the Clock: Temporal Alignment of
               Pretrained Language Models},
  booktitle = {Findings of the Association for Computational
               Linguistics: {ACL} 2024},
  pages     = {15010--15026},
  year      = {2024},
  address   = {Bangkok, Thailand},
  publisher = {Association for Computational Linguistics}
}

@inproceedings{guo2017calibration,
  author    = {Guo, Chuan and Pleiss, Geoff and
               Sun, Yu and Weinberger, Kilian Q.},
  title     = {On Calibration of Modern Neural Networks},
  booktitle = {Proceedings of the 34th International
               Conference on Machine Learning},
  pages     = {1321--1330},
  year      = {2017},
  publisher = {PMLR}
}

@preprint{wang2025categoryaware,
  author        = {Wang, Chen and Liu, Xunzhuo and Zhu, Yue and
                   Youssef, Alaa and Nagpurkar, Priya and Chen, Huamin},
  title         = {Category-Aware Semantic Caching for Heterogeneous {LLM} Workloads},
  year          = {2025},
  archivePrefix = {arXiv},
  eprint        = {2510.26835},
  note          = {Position paper}
}

@inproceedings{olston2008recrawl,
  author    = {Olston, Christopher and Pandey, Sandeep},
  title     = {Recrawl Scheduling Based on Information Longevity},
  booktitle = {Proceedings of the 17th International Conference on World Wide Web},
  series    = {WWW '08},
  year      = {2008},
  pages     = {437--446},
  publisher = {ACM},
  doi       = {10.1145/1367497.1367557},
}

\appendix

\section{Comparison with Related Systems}
\label{app:related}

Table~\ref{tab:related} summarizes how FreshCache differs from prior semantic caching, RAG acceleration, and temporal knowledge systems along six dimensions: freshness awareness, open-web applicability, multi-tier structure, explicit stale risk modeling, temporal decay modeling, and the presence of a temporal benchmark.

\begin{table*}[!t]
\centering
\caption{Comparison of FreshCache with related systems across key design dimensions. \checkmark~= fully addressed, $\sim$~= partially addressed, and $\times$~= not addressed.}
\label{tab:related}
\scriptsize
\renewcommand{\arraystretch}{1.15}
\setlength{\tabcolsep}{4pt}
\begin{tabular*}{\textwidth}{@{\extracolsep{\fill}}lcccccc@{}}
\toprule
\textbf{System} &
\textbf{\shortstack{Fresh.\\Aware}} &
\textbf{\shortstack{Web\\RAG}} &
\textbf{\shortstack{Multi-\\tier}} &
\textbf{\shortstack{Stale\\Risk}} &
\textbf{\shortstack{Temp.\\Decay}} &
\textbf{\shortstack{Temp.\\Bench.}} \\
\midrule
GPTCache~\cite{bang2023gptcache}              & $\times$ & $\times$    & $\times$ & $\times$    & \checkmark & $\times$ \\
SCALM~\cite{li2024scalm}                      & $\times$ & $\times$    & $\times$ & $\times$    & \checkmark & $\times$ \\
vCache~\cite{schroeder2025vcache}             & $\times$ & $\times$    & $\times$ & $\sim$      & \checkmark & $\times$ \\
CAG~\cite{chan2025cag}                        & $\times$ & $\times$    & $\times$ & $\times$    & $\times$   & $\times$ \\
RAGCache~\cite{jin2024ragcache}               & $\times$ & $\times$    & $\sim$   & $\times$    & $\times$   & $\times$ \\
CacheBlend~\cite{yao2025cacheblend}           & $\times$ & $\times$    & $\sim$   & $\times$    & $\times$   & $\times$ \\
FreshLLMs~\cite{vu2024freshllms}              & $\sim$   & \checkmark  & $\times$ & $\times$    & $\times$   & $\sim$ \\
DPR~\cite{karpukhin2020dpr}                   & $\times$ & $\times$    & $\times$ & $\times$    & \checkmark & $\times$ \\
ROME~\cite{meng2022rome}                      & $\sim$   & $\times$    & $\times$ & $\times$    & $\times$   & $\times$ \\
Time-Aware LM~\cite{dhingra2022time}          & $\sim$   & $\times$    & $\times$ & $\times$    & $\times$   & $\sim$ \\
RealTime QA~\cite{kasai2023realtime}          & $\sim$   & \checkmark  & $\times$ & $\times$    & $\times$   & \checkmark \\
Wang et al.~\cite{wang2025categoryaware}      & $\sim$   & $\times$    & $\times$ & $\times$    & $\sim$     & $\times$ \\
Web Evolution~\cite{fetterly2003evolution}    & $\sim$   & $\times$    & $\times$ & $\times$    & $\sim$     & $\sim$ \\
\midrule
\textbf{FreshCache (Ours)}                    & \checkmark & \checkmark & \checkmark & \checkmark & \checkmark & \checkmark \\
\bottomrule
\end{tabular*}
\end{table*}

\section{Adversarial Entity-Swap Evaluation}
\label{app:adversarial}

Table~\ref{tab:adversarial} lists the 20 adversarial query pairs referenced in Section~\ref{sec:tiers}, each with BGE-M3 cosine similarity at or above 0.40 but asking about different entities. Cosine similarity alone approves all 20 as L1 candidates; the entity match filter described in Section~\ref{sec:tiers} blocks 15 of them.

\begin{table*}[!t]
\centering
\caption{Adversarial entity-swap evaluation. Each pair has BGE-M3 cosine similarity $\ge 0.40$ but asks about different entities. Cosine similarity alone approves all 20 pairs; the entity filter blocks 15.}
\label{tab:adversarial}
\scriptsize
\renewcommand{\arraystretch}{1.12}
\setlength{\tabcolsep}{4pt}
\begin{tabularx}{\textwidth}{@{}>{\raggedright\arraybackslash}X
                                >{\raggedright\arraybackslash}X
                                cc@{}}
\toprule
\textbf{Query 1} & \textbf{Query 2} & \textbf{Sim.} & \textbf{Filter} \\
\midrule
Who did King Oedipus marry? & Which Beatle produced Life of Brian? & 0.948 & pass \\
Moon/Sun gravitational pull? & Which Beatle produced Life of Brian? & 0.946 & \textbf{block} \\
Academy Awards household rating? & Top goalscorer in Superettan? & 0.946 & \textbf{block} \\
Race time, W\"urth 400 winner? & President of Harvard Crimson? & 0.944 & pass \\
WWI Armistice rail car type? & Island between North/South Uist? & 0.940 & \textbf{block} \\
Georgia rugby head coach? & Big Ten women's basketball venue? & 0.925 & pass \\
Australian egg-laying mammal? & Amateur Ryder Cup equivalent? & 0.915 & \textbf{block} \\
Governor of Turks \& Caicos? & Kerala Premier League champion? & 0.901 & pass \\
KFC Uerdingen 05 division? & CMA Male Vocalist winner? & 0.893 & \textbf{block} \\
Miss South Dakota hometown? & Rally Guanajuato Mexico winner? & 0.880 & \textbf{block} \\
English motorway services? & Zodiac sign of the ram? & 0.871 & \textbf{block} \\
Arnold Clavio latest show? & California 47th district rep? & 0.857 & \textbf{block} \\
Lusitania sinking year? & First data-processing machine? & 0.836 & \textbf{block} \\
Miss Michigan USA city? & USA Gymnastics Championships venue? & 0.831 & \textbf{block} \\
Vic Sotto latest show? & Latest Freeform series network? & 0.828 & pass \\
Czech football champion? & Connecticut Comptroller hometown? & 0.824 & \textbf{block} \\
U.S. Figure Skating venue? & Guldpucken award recipient? & 0.795 & \textbf{block} \\
iPhone 16 price? & iPhone 16 Pro price? & 0.950 & \textbf{block} \\
CEO of OpenAI? & CEO of Google? & 0.920 & \textbf{block} \\
Champions League final score? & Europa League final score? & 0.910 & \textbf{block} \\
\midrule
\textbf{Total} & & \textbf{20 pairs} & \textbf{15 blocked} \\
\bottomrule
\end{tabularx}
\end{table*}

\section{Tier Multiplier Sensitivity}
\label{app:tiermult}

The tier multipliers introduced in Equation~\ref{eq:tier} are fixed constants without a closed-form fitting procedure. To confirm reported performance is not an artifact of this choice, we sweep $m_\text{answer} \in \{1.0, 1.25, 1.5, 2.0\}$ at $t=24$h while holding the other multipliers fixed. Search savings remain flat at 98.1\% across this range, and stale error varies only between 1.5\% and 1.7\%, indicating FreshCache's reuse decisions are governed primarily by the calibrated half-lives and error budgets rather than by the specific value of $m_\text{answer}$. We report $m_\text{answer} = 1.5$ as the deployed value, motivated by the answer tier's higher staleness sensitivity, but the system is robust to this choice within the tested range.

\section{Implementation Hyperparameters}
\label{app:impl}

Table~\ref{tab:impl} lists the hyperparameters used across all experiments in Section~\ref{sec:impl}.

\begin{table}[!t]
\centering
\caption{Key hyperparameters. Latency values marked $\dagger$ are simulation model constants; Appendix~\ref{app:liveweb} reports measured values from a 149-URL live-web microbenchmark.}
\label{tab:impl}
\small
\renewcommand{\arraystretch}{1.12}
\setlength{\tabcolsep}{4pt}
\begin{tabularx}{\columnwidth}{@{}>{\raggedright\arraybackslash}X l@{}}
\toprule
\textbf{Parameter} & \textbf{Value} \\
\midrule
Similarity encoder & BAAI/bge-m3 \\
L1 similarity threshold & 0.40 \\
L2 similarity threshold & 0.35 \\
L1 error budget $\varepsilon_{\mathrm{L1}}$ & 0.10 \\
L2 error budget $\varepsilon_{\mathrm{L2}}$ & 0.20 \\
L3 error budget $\varepsilon_{\mathrm{L3}}$ & 0.35 \\
Tier multipliers $(m_{\mathrm{L1}}, m_{\mathrm{L2}}, m_{\mathrm{L3}})$ & 1.5, 1.2, 1.0 \\
MLP hidden layers & [64, 32, 16] \\
MLP dropout & 0.30 \\
Temperature $T$ & 1.4672 \\
Search API latency & 500ms$^\dagger$ \\
Web fetch latency & 800ms$^\dagger$ \\
Conditional GET latency & 150ms$^\dagger$ \\
LLM generation latency & 2{,}000ms \\
\bottomrule
\end{tabularx}
\end{table}

\section{Results at $t = 1$ Hour}
\label{app:t1h}

Table~\ref{tab:main1} reports the same metrics as Table~\ref{tab:main24} at the one hour window, referenced in Section~\ref{sec:mainresults}.

\begin{table}[!t]
\centering
\caption{Results at $t = 1$h on all classes using BGE-M3 encoder
(31,201 queries, 477 stale URLs). S/q = search calls per query;
Stale = stale error rate; Saved = search savings. The primary deployable FreshCache result is shown in \textbf{bold}.
$\dagger$ = oracle upper bound, not deployable.}
\label{tab:main1}
\small
\renewcommand{\arraystretch}{1.12}
\setlength{\tabcolsep}{4pt}
\begin{tabular*}{\columnwidth}{@{\extracolsep{\fill}}lccc@{}}
\toprule
\textbf{Method} & \textbf{S/q} & \textbf{Stale} & \textbf{Saved} \\
\midrule
SemanticTTL      & 0.03 & 16.2\% & 97\% \\
DomainTTL        & 0.28 & 9.4\%  & 72\% \\
TempKeyTTL       & 0.12 & 9.6\%  & 88\% \\
vCache           & 0.52 & 4.4\%  & 48\% \\
SCALM            & 0.04 & 3.6\%  & 96\% \\
TieredTTL        & 0.01 & 16.0\% & 99\% \\
\midrule
FC\_NoCalib      & 0.02 & 8.2\%  & 98\% \\
FreshCache       & 0.02 & 2.6\%  & 98\% \\
\textbf{FC\_MLP} & \textbf{0.02} & \textbf{2.0\%} & \textbf{98\%} \\
\midrule
Oracle$^\dagger$ & 0.01 & \textit{0.0\%} & \textit{99\%} \\
\bottomrule
\end{tabular*}
\end{table}

\section{TTL Pareto Frontier Sweep}
\label{app:ttlsweep}

\begin{figure}[!t]
    \centering
    \includegraphics[width=\columnwidth]{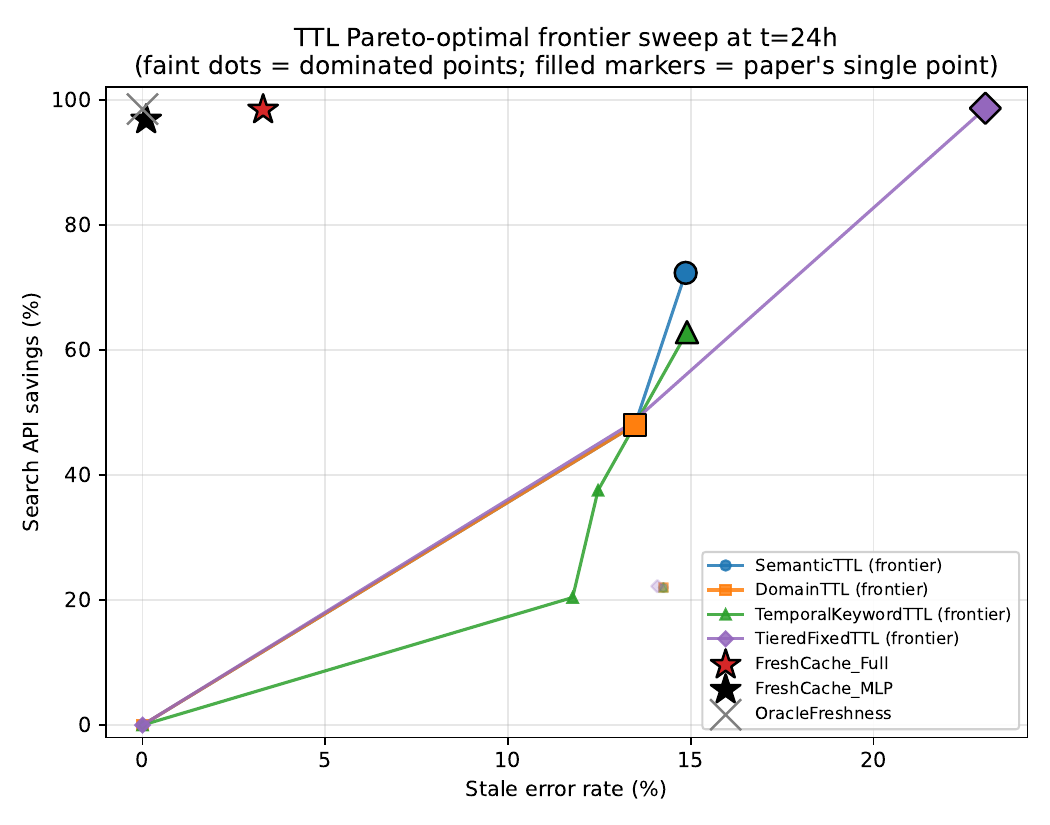}
    \caption{TTL Pareto frontier sweep at $t = 24$h using BGE-M3. TTL-based baselines are evaluated across eight scale factors, from 0.01$\times$ to 1.5$\times$ their default TTLs. Filled markers denote the default operating points from Table~\ref{tab:main24}, while faint dots denote dominated settings. FreshCache\_Full and FreshCache\_MLP are fixed reference points.}
    \label{fig:pareto_ttl}
\end{figure}

Figure~\ref{fig:pareto_ttl} sweeps each TTL-based baseline across eight scale factors, from 0.01$\times$ to 1.5$\times$ its default TTL, referenced in Section~\ref{sec:mainresults}. No configuration on any baseline frontier simultaneously achieves lower stale error and higher search savings than FreshCache\_Full's 3.3\% stale, 98.4\% saved operating point.

\section{Live-Web Microbenchmark}
\label{app:liveweb}

Referenced in Sections~\ref{sec:cget}, \ref{sec:liveweb_cise}, and \ref{sec:discussion}. We fetched 149 URLs directly from the live web, approximately 30 per freshness class, and issued a conditional GET for the subset with a stored ETag or Last-Modified header. Table~\ref{tab:liveweb} reports the aggregate results.

\begin{table}[!t]
\centering
\caption{Live-web microbenchmark on 149 URLs, with approximately 30 URLs per class. All latencies are in ms. Sim. = simulation constant; Meas. = measured value.}
\label{tab:liveweb}
\small
\renewcommand{\arraystretch}{1.12}
\setlength{\tabcolsep}{4pt}
\begin{tabularx}{\columnwidth}{@{}>{\raggedright\arraybackslash}Xrrr@{}}
\toprule
\textbf{Metric} & \textbf{Sim.} & \textbf{Meas.} & \textbf{Rel.} \\
\midrule
\multicolumn{4}{@{}l}{\textit{Cold fetch (n = 141 / 149)}} \\
Mean & 800 & 1{,}148 & $+$43.5\% \\
p50  & 800 & 915     & \checkmark \\
p90  & 800 & 2{,}170 & 2.7$\times$ \\
p95  & 800 & 2{,}771 & 3.5$\times$ \\
p99  & 800 & 4{,}573 & 5.7$\times$ \\
\midrule
\multicolumn{4}{@{}l}{\textit{Conditional GET (n = 53)}} \\
Mean     & 150 & 565    & 3.8$\times$ \\
304 rate & --- & 11.3\% & --- \\
\bottomrule
\end{tabularx}

\vspace{1mm}
\begin{minipage}{\columnwidth}
\scriptsize
Rel. = relative change compared with the simulation constant. The 304 rate shows that only 11.3\% of conditional GET requests returned 304; most remaining responses were full 200 responses.
\end{minipage}
\end{table}

Median cold-fetch latency of 915ms is close to the assumed 800ms, but at p95 real latency reaches 2,771ms, 3.5 times the simulated constant, meaning every L1 or L2 hit served to a request that would otherwise fall at that percentile saves nearly 2,800ms rather than 800ms. Per-class cold-fetch medians range from 761ms to 915ms with no statistically meaningful difference across classes, confirming the single 800ms constant is a reasonable class-agnostic approximation. Only 11.3\% of the 53 servers with caching headers honored the conditional request; the remaining 88.7\% returned a full 200 response regardless of the If-None-Match or If-Modified-Since header, and average conditional GET latency at 565ms left little advantage over a cold fetch. This means L3+ConditionalGet is more expensive to operate in production than the simulation suggests, and reinforces why the L1 and L2 tiers, which avoid search, fetch, and validation entirely, carry most of FreshCache's practical value.

\section{Cache-Induced Stale Error on External Benchmarks}
\label{app:cise}

Referenced in Sections~\ref{sec:perclass}, \ref{sec:liveweb_cise}, and \ref{sec:conclusion}. To measure cache-induced stale error directly, defined as a cache hit serving a wrong answer that a fresh retrieval would have answered correctly, we evaluate FreshCache on two external, publicly available QA benchmarks with verified answer-change labels, neither used to calibrate or train any FreshCache component. Freshness class for both benchmarks is assigned using the same lightweight classifier FreshCache uses at inference, not the benchmark's own change label, to avoid leaking the evaluation outcome into the reuse decision.

On 43,453 EvolvingQA question pairs (10,000 changed, 33,453 unchanged), FreshCache\_MLP reuses only 15.6\% of queries, reflecting that its features, calibrated on FreshCache-Bench's web snapshot domain volatility distribution, do not transfer cleanly to Wikipedia-style factual QA. Among cached queries, 29.2\% result in a cache-induced stale error, with overall answer accuracy at 95.5\%. On 43,456 DailyQA consecutive day answer pairs covering sports, finance, and current events, the domain shift is more severe: FreshCache\_MLP's hit rate falls to effectively 0\%, since nearly every query is treated as too risky under this benchmark's real URL volatility and evidence count features. The rule-based model remains functional in both cases, achieving a 25.1\% hit rate at 16.3\% cache-induced stale error on DailyQA.

Both results point to the same cause: the MLP's features describe web page and domain context learned from FreshCache-Bench's snapshot change distribution, not the real-world event timing that determines whether a specific fact changed on a specific day. This is a domain transfer limitation rather than an implementation defect, and it grows more pronounced as the target domain's volatility profile diverges from the training distribution. The rule-based model, calibrated only on aggregate per-class change rates, degrades more gracefully across both external domains and remains the recommended fallback when the learned model's training distribution does not match the deployment workload.

\section{Workload Characterization}
\label{app:workload}

\begin{figure}[!t]
    \centering
    \includegraphics[width=\columnwidth]{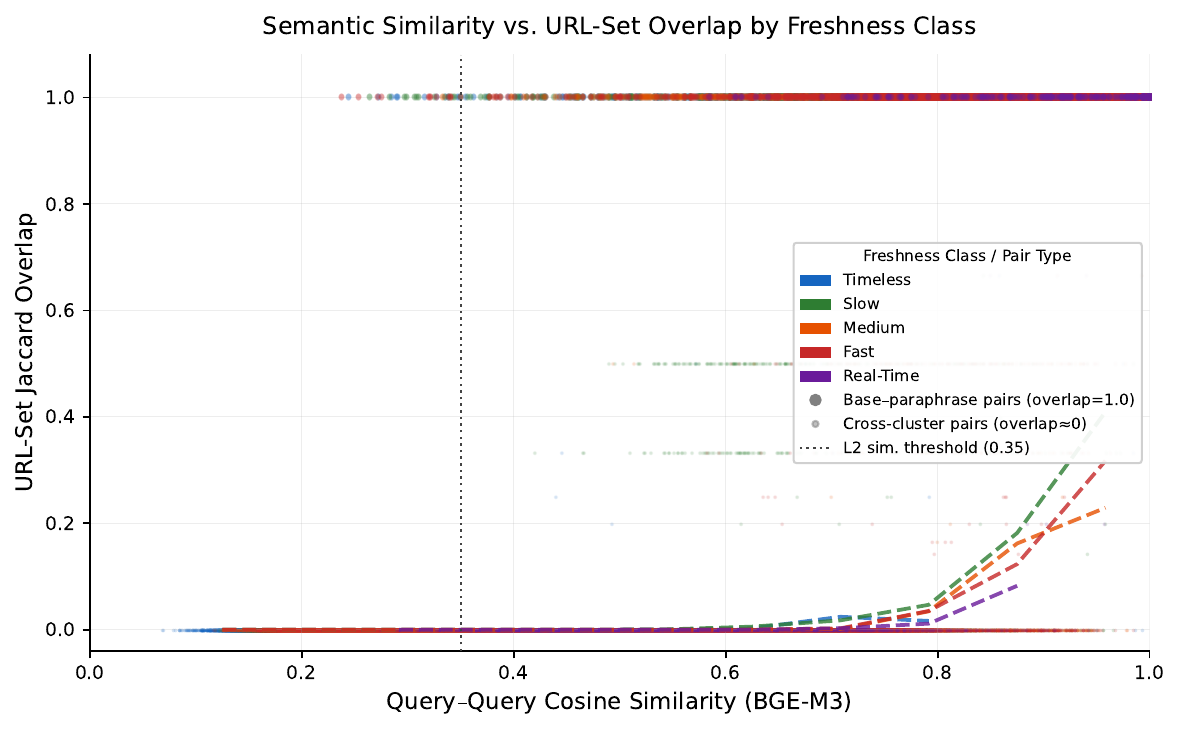}
    \caption{Workload characterization: cosine similarity between query pairs vs. Jaccard overlap of their top-returned URL sets, colored by freshness class. TIMELESS and SLOW pairs with high semantic similarity consistently share URL sets, empirically validating the L2 URL-list cache for stable classes. MEDIUM and FAST pairs show substantially lower URL overlap at equivalent similarity, explaining why L2 reuse requires a stricter risk gate and why the tier shift from L1 to L2 is selective rather than uniform.}
    \label{fig:workload}
\end{figure}

Referenced in Section~\ref{sec:tiershift}. Figure~\ref{fig:workload} plots query-pair cosine similarity against URL-set Jaccard overlap for base-to-paraphrase and cross-cluster pairs, colored by freshness class, validating the assumption behind the L2 tier that semantically similar queries tend to retrieve the same URLs.

TIMELESS and SLOW pairs with high semantic similarity reliably share URL sets, making L2 reuse both safe and effective for these classes. MEDIUM and FAST pairs show high semantic similarity but substantially lower URL overlap, which is precisely why the risk gate blocks L2 reuse for these classes at 24 hours and falls back to a fresh search.

\section{LLM-Judge Per-Class Breakdown}
\label{app:llmjudge}

Referenced in Section~\ref{sec:llmjudge}. Table~\ref{tab:llm_judge} breaks the 396 evaluated pairs down by freshness class.

\begin{table}[!t]
\centering
\caption{LLM-judge metric validation at $t = 24$h. For each URL-query pair with a hash-detected content change, a local Llama-3.2-3B-Instruct model judged whether the change would make the cached answer factually wrong. YES = answer would be wrong; NO = cosmetic or irrelevant change.}
\label{tab:llm_judge}
\small
\renewcommand{\arraystretch}{1.12}
\setlength{\tabcolsep}{4pt}
\begin{tabular*}{\columnwidth}{@{\extracolsep{\fill}}lrrrr@{}}
\toprule
\textbf{Class} & \textbf{Pairs} & \textbf{YES} & \textbf{NO} & \textbf{True Stale} \\
\midrule
TIMELESS   & 22  & 1   & 21  & 4.5\%  \\
SLOW       & 53  & 11  & 42  & 20.8\% \\
MEDIUM     & 104 & 21  & 83  & 20.2\% \\
FAST       & 126 & 42  & 84  & 33.3\% \\
REAL\_TIME & 91  & 61  & 30  & 67.0\% \\
\midrule
\textbf{ALL} 
           & \textbf{396} & \textbf{136} & \textbf{260} & \textbf{34.3\%} \\
\bottomrule
\end{tabular*}
\end{table}

The pattern across classes matches the freshness taxonomy: TIMELESS shows a very low answer-affecting change rate (4.5\%), since its pages occasionally refresh layout or metadata but rarely change the underlying facts at a 24-hour timescale, while FAST and REAL\_TIME change in ways that genuinely matter at 33.3\% and 67.0\%, consistent with treating these classes as high risk.

\section{Risk Gate Ablation, Full Per-Class Breakdown}
\label{app:riskgate}

Referenced in Section~\ref{sec:ablation_riskgate}. Table~\ref{tab:ablation_riskgate} reports the class-by-class comparison between SemanticTTL and rule-based FreshCache at both evaluation windows.

\begin{table*}
\centering
\caption{Risk gate ablation: SemanticTTL vs. rule-based FreshCache per freshness class using BGE-M3. Stale = stale error rate; Saved = search savings. $\ast$ = SemanticTTL TTL for FAST expired at 24h. $\Delta$ = absolute reduction in stale error from SemanticTTL to FreshCache.}
\label{tab:ablation_riskgate}
\scriptsize
\renewcommand{\arraystretch}{1.12}
\setlength{\tabcolsep}{4pt}
\begin{tabular*}{\textwidth}{@{\extracolsep{\fill}}lcccccccc@{}}
\toprule
\multirow{2}{*}{\textbf{Class}} 
& \multicolumn{4}{c}{$t = 1$h} 
& \multicolumn{4}{c}{$t = 24$h} \\
\cmidrule(lr){2-5} \cmidrule(lr){6-9}
& \textbf{SemTTL} & \textbf{FC} & \textbf{$\Delta$} & \textbf{Saved}
& \textbf{SemTTL} & \textbf{FC} & \textbf{$\Delta$} & \textbf{Saved} \\
\midrule
TIMELESS & 0.9\%  & 1.0\% & $-0.1$        & 98\% 
         & 2.6\%  & 1.5\% & \textbf{1.1}  & 98\% \\
SLOW     & 8.2\%  & 1.5\% & \textbf{6.7}  & 99\% 
         & 9.5\%  & 2.9\% & \textbf{6.6}  & 99\% \\
MEDIUM   & 4.7\%  & 4.1\% & 0.6           & 99\% 
         & 5.9\%  & 4.3\% & \textbf{1.6}  & 0\% \\
FAST     & 3.1\%  & 5.6\% & $-2.5$        & 0\%  
         & 0.0\%$^\ast$ & 5.4\% & ---      & 0\% \\
\midrule
\textbf{ALL} 
         & 16.2\% & 2.6\% & \textbf{13.6} & 98\% 
         & 14.9\% & 3.3\% & \textbf{11.6} & 98\% \\
\bottomrule
\end{tabular*}

\vspace{1mm}
\begin{minipage}{\textwidth}
\scriptsize
SemTTL = SemanticTTL; FC = FreshCache. All results use the BGE-M3 encoder.
\end{minipage}
\end{table*}

\end{document}